# Diminishing Uncertainty within the Training Pool:
# Active Learning for Biomedical Image Segmentation

Vishwesh Nath, Dong Yang, Bennett A. Landman, Daguang Xu, Holger R. Roth

NVIDIA, Bethesda, USA

Contact: vnath@nvidia.com, hroth@nvidia.com

## Abstract

*Active learning is a unique abstraction of machine learning techniques where the model/algorithm could guide users for annotation of a set of data points that would be beneficial to the model, unlike passive machine learning. The primary advantage being that active learning frameworks select data points that can accelerate the learning process of a model and can reduce the amount of data needed to achieve full accuracy as compared to a model trained on a randomly acquired data set. Multiple frameworks for active learning combined with deep learning have been proposed, and the majority of them are dedicated to classification tasks. Herein, we explore active learning for the task of segmentation of medical imaging data sets. We investigate our proposed framework using two datasets: 1.) MRI scans of the hippocampus, 2.) CT scans of pancreas and tumors. This work presents a query-by-committee approach for active learning where a joint optimizer is used for the committee. At the same time, we propose three new strategies for active learning: 1.) increasing frequency of uncertain data to bias the training data set; 2.) Using mutual information among the input images as a regularizer for acquisition to ensure diversity in the training dataset; 3.) adaptation of Dice log-likelihood for Stein variational gradient descent (SVGD). The results indicate an improvement in terms of data reduction by achieving full accuracy while only using 22.69 % and 48.85 % of the available data for each dataset, respectively.*

## 1. Introduction

Recent times have seen the emergence of deep learning at an exponentially growing rate. There are multiple standard deep learning network architectures that have been established such as Inception [37], ResNet-50 [16], U-Net [32], and V-Net [29]. While deep learning has shown exceptional performance and accuracy for classification, regression, segmentation, and natural language processing related tasks, it has been limited by its hunger for large labeled datasets, which are difficult to collect, especially in the medical imaging setting. A recent stipulation has been the usage of targeted data selection techniques to enhance the performance of deep learning networks with limited amounts of data. These techniques typically refer to a concept of machine learning, commonly known as active learning [33, 5].

Active learning, combined with deep learning, allows for the development of a framework in which the deep network architecture is coupled with classical techniques for selection of data points (Fig. 1). Active selection of data for a model promises to lead to faster convergence, increased performance with fewer data, and improved robustness due to its targeted selection of data points that are characterizable as outliers or hard examples. The inception of active learning goes back to 1988 when the first initial work was characterized using queries and concept learning [1]. There are various frameworks for active learning, such as uncertainty sampling, query-by-committee, variance reduction, or expected model change [33]. One of the most widely used frameworks is uncertainty sampling, where a variety of approaches have been proposed [12, 3, 35, 30, 11, 43, 44]. Active learning acquisition schemes require heavy computation (unlabeled pool of data is usually relatively large) in addition to the computational burden of training deep networks, which often have a large number of parameters (Inception V3 has > 23 million parameters). The computations for acquisition schemes on large unlabelled data have become tractable for linear acquisition schemes but not when framed as a non-linear optimization objective [10]. The primary areas of applications for deep active learning are mainly concerned with the reduction of labeling cost when acquiring training data for deep learning [5]. Primary applications for active learning are natural language processing, speech recognition, classification, and in particular, medical imaging. For example, a skilled radiologist typically requires several hours to outline the abdominal organs of one 3D CT image [31].

We would also like to bring attention to the fact that active learning in general is a challenging problem, and working solutions are highly dataset dependent [15]. Few works acknowledge that random acquisition is a surprisingly difficult baseline to beat. For example, [15] illustrated the dataset dependency for the success of active learning over random acquisition of data points in their experimental results. In the context of medical imaging, active learning becomes more challenging due to the inherently complex nature of the data.

In this work, we explore the benefits of active learning for deep learning, specifically for medical image segmentation tasks. We explore the acquisition schemes based on uncertainties estimated by a query-by-committee framework. A query-by-committee framework consists of multiple models with similar architectures that are independently trained to estimate these uncertainties. To the best of our knowledge, previous acquisition schemes have been based only on the unlabeled pool of data and do not utilize the information presented in the training pool of already labeled images in any manner. Our approach aims to use already labeled images more efficiently. This is achieved by using active learning to increase the frequency of uncertain cases in the training pool of data to decrease the overall model uncertainty. While the frequency of uncertain data points is increased, we also propose to regularize the data of the training pool with the usage of mutual information between the training pool and the unlabeled pool of data to guarantee a varied enough training set. To efficiently implement the query-by-committee framework, we introduce a novel adaptation of Stein variational gradient descent (SVGD) [25]. SVGD is a joint optimization technique for an ensemble of models where each model is referred to as a particle. The particles are jointly updated at each step and weighted by a radial basis function (RBF) kernel. SVGD was initially proposed with the usage of log-likelihood and cross-entropy. Here, we adapt SVGD for segmentation tasks by proposing a Dice loss based log-likelihood definition and utilize each particle's predictions for uncertainty estimation in medical image segmentation tasks.

## 2. Related Work

### 2.1. Early Work

Active learning has a broad range of literature from the machine learning domain where the early work was based on acquisition functions using the variance of predictions [7, 39]. These prior methods were proposed in conjunction with classical machine learning techniques, such as linear regression, support vector machines and so forth. While the acquisition functions are a straight-forward approach for classification datasets. Similar approaches did not work well for biomedical image segmentation, and hence more complex acquisition schemes have been proposed [40]. Recent advancements proposed to use more complex acquisition functions combining epistemic & aleatoric uncertainty [41], Fisher information [35] and clustering techniques [47].

Early approaches for active learning in image segmentation were explored using support vector machines and with acquisition functions combining multiple classical measures such as entropy, region and smoothness of segmentation [40, 26]. However, approaching semantic segmentation using classical methods has almost become defunct and deep learning has become such a powerful tool for semantic segmentation that it is dominating the field. The same can be said for active deep learning frameworks.

### 2.2. Uncertainty Estimation of Models

There are three well-known ways to approximate the model uncertainty with deep neural networks: 1.) Monte-Carlo based sampling using attached dropout at different layers which are enabled during inference to generate uncertainty [11]; 2.) Bayesian neural networks where each parameter is characterized as a distribution. Hence the final predictions have a variance estimation directly [12, 4]; 3.) Using ensemble-based methods where the collection of models was utilized and the variance between the final predicted outcomes can serve as the uncertainty [6, 43]. Dropout-based Monte-Carlo sampling is a light-weighted model however it requires manual tuning of dropout hyper-parameters [36] which becomes an overhead for the user of active learning algorithms. Also, it should be noted that in practice dropout-based and Bayesian neural network uncertainty estimation do not work as well as ensemble-based models in practice [3, 24, 6]. Based on the conclusions of prior works, we propose an approach based on a *jointly-trained* ensemble of models.

### 2.3. Acquisition Schemes

A recent work combined the model uncertainty with representativeness of image similarity [43], where an ensemble of fully convolutional neural networks (FCN) were used for uncertainty estimation. The uncertainty was posed as a maximum cover set problem for optimal selection of data for active learning. The approach was limited by validation only on 2D imaging data [43]. Ensemble methods have also been used in conjunction with 3D patch-based FCN using a knapsack acquisition scheme to reach a sufficient budget [23]. The method was validated on a large dataset of 1247 CT scans for intracranial hemorrhage detection. Other approaches have been designed for active learning by using conditional generative adversarial networks (cGAN) [19] which synthesize data for an active learning framework utilizing a Bayesian neural network [4]. The approach was validated on a 2D chest radiograph dataset [28]. Patch-based convolutional neural networks (CNN) have also been coupled with Fisher information for active learning frameworks and improvements over epistemic entropy has been shown [35]. While current active learning in biomedical segmentation has been focused towards different acquisition schemes for annotation of data by the oracle (expert human annotator), other domains have also investigated reinforcement learning and proactive learning for active learning [9]. The most generic acquisition functions for multi-class segmentation are typically utilizing entropy-based uncertainty measures [30, 35, 6, 40, 5].

To summarize, multiple active learning frameworks with various acquisition schemes have been proposed. However, none shows a clear advantage over the other. Quite often the best-working acquisition scheme might be problem dependent. One thing is clear though, a successful active learning framework needs to have an implicit robust way of estimating the model uncertainty in the training pool.

### 2.4. Fine-tuning vs. from Scratch

Adding newly annotated data points changes the dynamics of the machine learning model as it needs to be generalizable to the newly added data points. Possible ways for model adaptation are either fine-tuning the existing model from the previous active iteration or re-training the entire model from scratch. There are numerous works that suggest that fine-tuning works well and it offers enhanced generalizability [38, 46]. It should be noted that the followers of fine-tuning also use various learning rate policies, freezing of specific model parameters, all of which bring additional overheads to users. At the same time, there are also multiple works that have suggested that fine-tuning is not applicable in all scenarios [2] and re-training from scratch is more beneficial. To the best of our knowledge, there is no work that suggests any disadvantages when training models from scratch apart from its computational cost. Hence, we choose to train networks from scratch for every active iteration.

## 3. Contributions

Most active learning approaches remove the selected data point from the unlabeled pool once it has been selected for annotation by the oracle. Active learning literature, in general, has avoided data duplication in the labeled/training pool, or ignored cases with high uncertainty (data rejection) [13, 45]. Recent work [20] has rightly suggested to develop methods that can deal with subject-level uncertainty and specifically data with high dimensionality and low sample size. Our work's primary contribution is the incorporation of data duplication into an active learning framework to combat highly uncertain and hard data points, instead of being neglected, which are utilized to: 1.) improve performance and 2.) reduce annotation cost. To regulate the amount of duplication, we utilize mutual information between data pairs of the training and unlabeled pool. We validate this contribution using a query-by-committee approach for active learning.

The following is an intuitive example of the motivation behind data duplication in training/labeled pool. If a dataset has three *inherent* classes $a, b, c$ and class $a, b$ have equivalent data points while class $c$ is sparse, the training of the machine learning model is likely to be biased towards classes where more data is presented (towards $a, b$ in the given case). However, if the hard-cases were duplicated from the class $c$, then it would lead to a more generalized model towards all classes. It should be noted that the inherent classes of the data mentioned here should not be confused with the different classes of labels, which is ground truth related information.

Ensemble models are generally used for query-by-committee active learning. However, it is often challenging to detect outlier models in an ensemble system. To overthrow this inhibition, we utilize a Bayesian inference algorithm for training ensembles where models consult each other so that they do not hinder their learning process. This algorithm is known as Stein variational gradient descent (SVGD) [25]. The SVGD algorithm was initially proposed with entropy-based likelihood. We adapt it towards a continuous Dice-based log-likelihood in order to adapt it to segmentation tasks. Furthermore, we adapt SVGD for active learning to infer uncertainty from the different model instances trained during SVGD.

In brief, our contributions can be summarized as below:

1. To the best of our knowledge, this is the first comprehensive study of multiple methods for active learning for medical image segmentation. We conducted an elaborate experimentation with realistic annotation query sizes for human annotators.

2. We explore the utility of hard-samples by increasing their frequency in the training pool of data to annotation cost while training effective models. At the same time, we limit the duplication of hard-cases by utilizing mutual information between the training pool and the unlabeled pool while achieving good uncertainty estimates.

3. We introduce SVGD with Dice log-likelihood as an optimizer naturally suited for query by committee active learning frameworks.

## 4. Proposed Method

### 4.1. Deep Learning Models

The nature of our method is not specific to a particular deep learning network architecture. Without loss of generality, we use a U-Net based architecture, which is popular for bio-medical segmentation [32]. Residual blocks were constructed per block of encoder and decoder structure of U-Net as in [29], as residual blocks, in general, are helpful for training and preventing overfitting [16]. We used 4 encoding layers and 3 decoding layers for the U-Net architecture. Initial filters were set to 8 for all data sets. Batch normalization and "relu" activation were used for each layer except the last layer which was activated by a "softmax" layer. An ensemble of U-Net models can be used to form a committee. Prior literature has shown the query-by-committee approach to work well for medical image segmentation with active learning [43, 3, 42, 8].

### 4.2. Stein Variational Gradient Descent (SVGD)

We utilize SVGD [25] as the query-by-committee approach used in our proposed active learning scheme. Despite SVGD being a variational inference algorithm, it has a deterministic update rule and can infer $M$ samples from a target distribution. The SVGD framework maintains $M$ copies of model parameters that are referred to as particles $\Theta = \{\theta^q\}_{q=1}^M$. This algorithm, in particular, has multiple useful properties that are applicable for active learning: 1.) SVGD has been shown to be a robust optimizer for training ensembles and it prevents convergence to outlier models by utilizing kernels. 2.) SVGD also ensures that each model finds a unique local optimum by utilizing repulsive forces in its update step (Eq. 1). At iteration $k$, each particle $\theta_k \in \Theta_k$ is updated using the step: $\theta_{k+1} \leftarrow \theta_k + \epsilon_k \phi(\theta_k)$ where,

$$\phi(\theta_k) = \frac{1}{M} \sum_{j=1}^{M} [r(\theta_k^j, \theta_k) \nabla_{\theta_k^j} log\ p(\theta_k^j) + \nabla_{\theta_k^j} r(\theta_k^j, \theta_k)].$$
(1)

Here, $\epsilon_k$ is the step-size and $r(\theta, \theta^{'})$ is a positive-definite kernel. The kernel is formed using RBF. The crucial aspect here is that each particle consults other particles to determine its course of gradient descent while ensuring a repulsive force exists between two particles (last term in ((1))).

The original SVGD formulation utilizes entropy-based log-likelihood. However, semantic segmentation for medical imaging often utilizes Dice-based continuous loss [29] over the traditional cross-entropy loss. This is due to the sparsity of semantic labels and class imbalance issue existing in labels. There are multiple known variants of Dice-based losses that can be combined with boundary based alternatives [22],[21]. For simplicity we concentrate on Dice based losses only without exploring boundary-enhancing alternatives. Hence, we adapt Dice log-likelihood which is defined by $log(\mathcal{L}_{Dice})$ with the Dice loss $\mathcal{L}_{Dice}$ defined as:

$$\mathcal{L}_{Dice}(y, \hat{y}) = 1 - \frac{2 \sum_{i=1}^n y_i \hat{y}_i}{\sum_{i=1}^n y_i^2 + \sum_{i=1}^n \hat{y}_i^2}.$$
(2)

Here, $y$ and $\hat{y}$ represent the ground truth and the prediction respectively. The Dice loss is preferable for medical imaging as the cross-entropy loss needs to be weighted to deal with the data imbalance in medical image segmentation tasks. It can be observed that a SVGD-based model drives the particles towards convergence with a lower standard deviation as compared to an ensemble-based model. See Fig. 2.

### 4.3. Active learning framework

As it would be cumbersome to have an actual human for annotation of data in experiments, we use a fully labeled dataset splitting into a training pool $\mathcal{T}$ and an unlabeled pool $\mathcal{U}$ consisting of $m$ and $n$ samples respectively. Acquisition functions $\mathcal{A}$ are used after inference on the unlabeled pool $\mathcal{U}$ to obtain uncertainty estimates (Fig. 1). We use entropy-based epistemic or model uncertainty defined as:

$$\mathcal{H}(x_i^{\mathcal{U}}) = -\sum_{q=0}^{Q} \sum_{c=0}^{c} p(y_i = c|x_i^{\mathcal{U}}) log\ p(y_i = c|x_i^{\mathcal{U}})_q,\ (3)$$

where $y_i$ is the probability estimated for a class $c$ given an input sample $x_i$ belonging to the unlabeled pool $\mathcal{U}$. The entropy measure is calculated per particle $q$. To obtain a score, we use the sum across all the 3D voxels in the 3D uncertainty map $\mathcal{H}$. Each step of algorithm is further elaborated in the active learning algorithm (Algorithm 39).

The second acquisition function we use is a combination of epistemic uncertainty and a constraint of mutual information $\mathcal{MI}(x_i^{\mathcal{T}}, x_j^{\mathcal{U}})$ between $\mathcal{U}$ and $\mathcal{T}$ to ensure diversity for the training pool $\mathcal{T}$ defined as:

$$\mathcal{MI} = \sum_{x_i^{\mathcal{T}}} \sum_{x_j^{\mathcal{U}}} P(x_i^{\mathcal{T}}, x_j^{\mathcal{U}}) log \frac{P(x_i^{\mathcal{T}}, x_j^{\mathcal{U}})}{P(x_i^{\mathcal{T}})P(x_j^{\mathcal{U}})}. \quad (4)$$

The mutual information is computed for pairs of the input images and the images from the unlabeled pool (please note that no label information is being utilized). Here, $x_j^{\mathcal{U}}$ is a data point from the unlabeled pool while $x_i^{\mathcal{T}}$ a data point from training pool. $P(x_i^{\mathcal{T}}, x_j^{\mathcal{U}})$ is the joint probability and $P(x_i^{\mathcal{T}}), P(x_j^{\mathcal{U}})$ are marginal probabilities of the two images with respect to the joint histogram of the two images. The mutual information matrix will be of dimensions $(m, n)$. The intuition being that if duplicate samples are being picked the mutual information can act as a regularization term in the acquisition function. Before combining the two measures, epistemic uncertainty is normalized using min-max normalization. The mutual information matrix is reduced to a vector by taking the mean row-wise (across $m$) and then min-max normalized.

Thereafter, the final score is calculated by subtracting the mutual information $\mathcal{MI}$ from the entropy score $\mathcal{H}$ as lower mutual information depicts higher diversity as defined by

$$\text{Score} = \alpha(\mathcal{H}(x_i^{\mathcal{U}})) - \frac{1}{\alpha}(\mathcal{MI}(x_i^{\mathcal{U}}, \mathcal{T})). \quad (5)$$

The terms are weighted by hyper-parameter $\alpha$ for further tuning. Our baseline acquisition function is the random acquisition of data points from the unlabeled pool. The three variants of acquisition functions were tested with both cases (0,1) of data delete $\mathcal{D}$ flag leading to 6 variants in total. The data delete option is to purely determine whether to remove a data point from the unlabeled pool or not once it has been actively selected.

If the data point was already annotated, then the oracle does not need to label it again. Thus, it saves the oracle (annotator) the effort, and a data point may be re-utilized as a duplicate in the updated training pool $\mathcal{T}$. In other words, the method is not generating different annotations through a noisy oracle. It is directly

---

**Algorithm 1:** Active Learning Algorithm.

**1 Input:**
**2** $T = \{(x_i^T, y_i^T) \dots (x_m^T, y_m^T)\}\ i \in [1, m]$
**3** $U = \{(x_j^T, y_j^U) \dots (x_n^U, y_n^U)\}\ j \in [1, n]$
**4** $D$ : Delete Flag
**5** $UseMI$: Mutual Info Flag
**6** $q$: Queries per active iteration
**7 Output:**
**8** $M_t$ : Trained model at iteration $t$
**9** $L$ : Labeled Predictions
**10 Initialize:**
**11** $M_{t=0}$ : pre-trained model on $T$
**12** $t \leftarrow 1$
**13 repeat**
**14** | each $(x_j^U)\ \in\ U$
**15 until for do**
**16 end for**
**17** ;
**18** $H(x_j^U)$ Eq. (3)
**19 if** $Use\ MI$ **then**
**20** | **for** $each\ (x_i^T)\ \in\ T$ **do**
**21** | | $MI(x_j^U, x_i^T)$ Eq. (4)
**22** | **end for**
**23 else**
**24** | $excludeMI$
**25 end if**
**26 Calculate** Score Eq. (5)
**27 Sort** $Score$
**28 Extract** q queries with highest to $ScoreQ$
**29 if** $Q\ is\ Unique$ **then**
**30** | **Oracle** annotates $Q$
**31 end if**
**32 Update** $T \leftarrow T \cup Q$
**33 if** $D\ is\ True$ **then**
**34** | Delete $Q$ from $U$
**35 end if**
**36 Clear** all queries from $Q$
**37 Train** $M_t$ updated $T$ with SVGD Eq. (1)
**38** $t \leftarrow t + 1$
**39 Until** Desired Performance

---

utilizing already annotated data. There could be value in multiple annotations from multiple oracles to further study variations between annotators, but that is beyond the scope of this work. In medical imaging applications, the cost of annotating an image is relatively much higher than that of other annotation tasks, and duplication can lead to a direct reduction of annotation cost.

# 5. Experiments

## 5.1. Data Sets

We use data from medical segmentation decathlon (MSD) 2018 in part [34]. Two datasets were used from the MSD challenge: 1.) Pancreas and tumor segmentation based on 3D computed tomography (CT) volumes 2.) Hippocampus segmentation of two different regions of interest based on T1-weighted magnetic resonance imaging (MRI).

The hippocampus data set was kept at the original resolution and only normalization was applied as a pre-processing step. The clip range used was 0 to 2048. The total number of labeled samples available was 263, divided into initial training pool: 10, unlabeled pool: 153, validation: 50 and testing: 50 subjects.

The pancreas data set was down-sampled to 4.0 $mm$ isotropic resolution as pre-processing for rapid experimentation and the data was clipped in the range -87 to 199 (Hounsfield units) and normalized to $[0, 1]$. The total amount of labeled samples available was 281, a random initial training pool was selected for model initialization: 20, unlabeled pool: 201, validation: 30 and testing: 30 subjects.

It should be noted that the experiments are being conducted with all pre-labeled data from [34]. The unlabeled pool $\mathcal{U}$ of data was randomly selected and no ground truth information was utilized from the unlabeled pool for the studied active learning methods.

## 5.2. SVGD versus Ensemble

Both SVGD and ensemble based models were initialized with a fixed random seed and executed on a random selection of 30 data points from the hippocampus dataset. Validation set was used the same as all other experiments. A fixed learning rate of 0.001, and batch size of 8 were utilized for both models.

## 5.3. Active Learning

Each data set was executed for 40 active iterations. For every active iteration, $\mathcal{Q}$ queries selected from unlabeled pool $\mathcal{U}$ were annotated by oracle and then added to the training pool $\mathcal{T}$. $\mathcal{Q}$ were set to 5 and 1 for pancreas and hippocampus data sets respectively. Instead of using a consistent number of epochs we used a consistent number of training steps per each active iteration, as the training pool $\mathcal{T}$ grows in size with the addition of $\mathcal{Q}$. Hence, each active iteration was trained for a fixed number of steps which were determined by training on the full data set. The number of steps determined for pancreas and hippocampus is 10000 and 1500 respectively. For each data set, the active learning framework was repeated for 5 different seeds. The seeds were chosen randomly. The objective is to evaluate at the final outcome of 39 active iterations if a relative performance can be achieved with the training performed with the entire dataset.

## 5.4. Active Iteration Parameters

During each active iteration, a model is trained from scratch at a fixed random seed to ensure consistent initialization for the network. The number of particles for all experiments was fixed at 5. We used a constant learning rate per data set pancreas: 0.0004, hippocampus: 0.001. A batch size of 8 was used for both data sets. This means that 8 patches were selected for pancreas, however, for hippocampus 8 volumes were selected as all volumes were of the same size of $64 \times 64 \times 64$. The hyper-parameter $\alpha$ was set to 1 for Eq. (5).

For pancreas, a data set the 3D volumes were segregated into cubic patches for training. A consistent patch size of $48 \times 48 \times 48$ was used. For inference, a scanning-window technique was used at consistent strides of 36. The patches were selected for training on-the-fly using a positive/negative ratio of 1:1 where a positive patch ensured the existence of foreground in the patch and a negative patch is a random ROI cropped from the entire 3D volume.

The loss was purely evaluated on all classes of the data set for pancreas and background was excluded for hippocampus. Strictly, no data augmentation techniques were applied.

## 5.5. Ablation Studies

*Maximum-Cover Experiment:* To evaluate the effectiveness of the two proposed contributions of SVGD based models and data duplication in the training pool we tested with a prior acquisition approach for representativeness using maximum cover [43]. A total of 8 different variants were tested for the hippocampus dataset. Entropy and variance based uncertainty estimation were combined with maximum cover and then varied with deleting and not deleting data points (increasing frequency of hard-samples) from the unlabeled pool. The four acquisition schemes were tested for ensemble based models and SVGD based jointly driven ensemble models. These experiments were executed for 40 active iterations on a single fixed random seed for all 8 variants.

*Jenson-Shannon Divergence Experiment:* The Jenson-Shannon divergence (JSD) is used as a baseline acquisition function based on prior work [29]. The JSD acquisition was evaluated using combinations of maximum cover, and in variations of ensemble and SVGD as query-by-committee. This additional experiment was repeated with 5 different seed initializations for both hippocampus and pancreas dataset. The objective is to evaluate the utility of the existing baseline of ensemble based methods along with JSD, also incrementally testing the proposed methodologies of SVGD and increasing frequency of hard-samples (NoDelete).

A total of 8 active learning methodologies were studied using the two proposed strategies (SVGD, increasing frequency of hard-samples) and one existing strategy (Ensemble) with acquisition function of Jenson-Shannon divergence. The query size was set at 1 for both pancreas and hippocampus. The number of active iterations were set at 40. The learning rate, batch-size and optimizer were kept consistent as with all prior experiments as stated in Section V, D. Active Iteration Parameters.

### 5.6. Evaluation

The validation set was evaluated every $2^{nd}$ epoch. Dice-Sorensen Coefficient (DSC) was computed between the ground truth and inferred prediction for every 3D volume in the validation set. The validation accuracy is represented by the mean of all Dice scores of all 3D volumes of all classes per data set. A Wilcoxon signed-rank test was conducted for all pairs of methods on test and validation sets for both data sets, using the best active iteration model chosen based on the validation dataset. For both Table I and II, the asterisk indicates if the results were found to be significant ($p < 0.05$) by using Wilcoxon signed rank test. The asterisk was marked only if the result across all validation samples for a method were found to be significantly different from all other methods.

### 5.7. Implementation

This framework was implemented using TensorFlow v1.13 and utilized the Clara Train SDK[1]. Total of 60 NVIDIA Tesla V100 16GB GPUs were used in parallel for computation. Training time per active iteration for hippocampus and pancreas dataset were 45 minutes and 3 hours respectively.

## 6. Results

### 6.1. SVGD vs Ensemble

It can be observed that SVGD has a faster and smoother overall convergence (right most plot Fig. 2). Both models overfit the training data. Yet, the ensemble has a much higher standard deviation towards convergence at approximately 400 steps. It should also be noted that the models in the ensemble setup have a wider spread as compared to SVGD particles which strive for a consistent convergence while maintaining a certain distance in between each particle. This observation can be attributed to the repulsive force term in Eq. (1).

---

[1] https://developer.nvidia.com/clara

### 6.2. Hippocampus

Comparing mean Dice's score of all volumes in the validation set per active iteration for all methods Fig. 3 (Left), we can observe that random acquisition methods show a wider spread of variance for the first 15 active iterations. Relatively other methods show a restricted spread and overall higher Dice's scores. Higher than 15 active iterations show convergence towards the full performance of SVGD on the entire dataset. There is high variance for all methods while nearing convergence to full performance. In Fig. 3 (Right), it can be observed that the proposed methods of epistemic uncertainty combined with MI without deleting data points from the unlabeled pool attain an equivalent or higher Dice's scores as compared to methods where data was deleted. Notably, much less data is utilized by methods where data were not deleted from the unlabeled pool.

Tab. 1 presents a summary of the best active iteration model selected based on the validation scores for each method for hippocampus dataset. The mean validation scores are in the ballpark of the performance of SVGD on the full dataset. The maximum Dice for all the methods exceeds the performance of full dataset. Due to active selection the inherent classes of the training pool are balanced which leads to a higher performance. The same can be observed for the test set Dice scores. The epistemic uncertainty combined with MI without deleting data uses the least amount of data from the unlabeled pool with 22.69%.

Qualitatively, Fig 4 shows the efficiency of different methods for decreasing uncertainty in the following order: epistemic + MI with delete > epistemic no delete > epistemic with delete > epistemic + MI no delete. Specifically, we can observe a reduction of uncertainty on the difficult boundary regions of the segmented hippocampus head and body.

On observation of representative segmentation for different active learning methods (Fig. 5) specifically for a single class of probability it can be noticed that all methods perform well and are qualitatively similar. However, the proposed method Epistemic+MI+NoDelete utilizes a lesser amount of data to achieve a similar result (Tab. 1).

| Hippocampus Head Body and Tail Segmentation | | | |
|---|---|---|---|
| Acquisition Function | Val Mean Dice | Test Mean Dice | % Data |
| Epistemic Delete | 0.7246 ± 0.0135* | 0.7051 ± 0.0211 | 30.06 % |
| Epistemic+MI Delete | 0.7235 ± 0.0179* | 0.7014 ± 0.0142 | 30.06 % |
| Random No Delete | 0.7180 ± 0.0085* | 0.6985 ± 0.0116 | 28.22 % |
| Epistemic No Delete | 0.7211 ± 0.0110* | 0.6931 ± 0.0108 | 26.38 % |
| *Epistemic+MI No Delete* | **0.7241** ± 0.0176* | 0.7135 ± 0.0274* | **22.69** % |
| Random Delete baseline | 0.7039 ± 0.0058* | 0.6805 ± 0.0124 | 20.00 % |
| Random Delete baseline | 0.7194 ± 0.0125* | 0.7107 ± 0.0190 | 30.06 % |
| Full Dataset SVGD | 0.7257 | 0.7093 | 100 % |
| Full Dataset Ensemble | 0.7222 | 0.7031 | 100 % |

Table 1: Summarized Dice's scores along with % of data used for training from hippocampus datasets. The scores are based on the best active iteration model that was selected using highest mean Dice's score of the validation set. The asterisk describes statistical significance across pairs of methods using Wilcoxon signed rank test. Refer Section V, F. Evaluation for more details.

| Pancreas and Tumor Segmentation | | | |
|---|---|---|---|
| Acquisition Function | Val Mean Dice | Test Mean Dice | % Data |
| Random Delete | 0.4323 ± 0.0167 | 0.3698 ± 0.0156* | 97.01 % |
| Epistemic Delete | 0.4358 ± 0.0089 | 0.3658 ± 0.0115 | 89.55 % |
| Epistemic+MI Delete | 0.4309 ± 0.0085* | 0.3679 ± 0.0069* | 97.01 % |
| Random No Delete | 0.4175 ± 0.0244* | 0.3572 ± 0.0096 | 71.74 % |
| Epistemic No Delete | 0.3375 ± 0.0124* | 0.3153 ± 0.0292* | 21.19 % |
| *Epistemic+MI No Delete* | **0.3874** ± 0.0130* | 0.3358 ± 0.0102 | **48.85** % |
| Random baseline | 0.3244 ± 0.0084* | 0.2896 ± 0.0108 | 20.00 % |
| Random baseline | 0.3673 ± 0.0120* | 0.3424 ± 0.0197 | 40.00 % |
| Random baseline | 0.3739 ± 0.0062* | 0.3380 ± 0.0202 | 50.00 % |
| Full Dataset SVGD | 0.4230 | 0.3916 | 100 % |
| Full Dataset Ensemble | 0.4017 | 0.3624 | 100 % |

Table 2: Summarized Dice's score scores along with % of data used for training from pancreas dataset. The scores are based on the best active iteration model that was selected using highest mean Dice's score of the validation set. The asterisk describes statistical significance across pairs of methods using Wilcoxon signed rank test. Refer Section V, F. Evaluation for more details.

### 6.3. Hippocampus Ablation Experiment

It is evident that SVGD offers a better performance and a more reliable uncertainty estimation as compared to ensemble based methods when tested with other acquisition functions such as maximum cover [43] based (Fig. 6). It should also be noted that acquisition functions where data is not being deleted behave similarly and hence it can be suggestive that increasing frequency of hard-cases in the training pool is data-driven. It can be observed that for hippocampus dataset that SVGD allows for a faster convergence rate for all combinations (Fig. 7). For combinations where frequency of hard-samples are increased (NODelete), it can be observed that either a higher or similar performance can be attained with usage of lesser unique data points in the training pool.

### 6.4. Pancreas

Comparing to mean Dice's scores of all volumes in the validation set for pancreas data set Fig. 8 (left), we can observe consistent variance when different methods are inter-compared. It should be noted that the proposed no delete methods reach a plateau after 20 active iterations. The proposed delete methods achieve higher scores till 20 active iterations after which random delete Dice's score reaches similar performance. Towards convergence, the delete methods match the full baseline performance however methods with no delete from the unlabeled pool do not, see Fig. 8 (right). We can observe that no delete methods utilize lesser unique data points relatively as compared to delete methods.

Tab. 2 presents a summary of the best active iteration model selected based on the validation scores for each method for pancreas dataset. The mean validation Dice's scores are in the ballpark of the performance of SVGD on the full dataset except for epistemic with no delete from the unlabeled pool. The maximum Dice's score for all methods except epistemic with no delete attains similar or equivalent performance of full dataset. The same can be observed for the test set Dice's scores. The epistemic uncertainty without deleting data uses 21.19 % of data and when combined with MI uses 48.85 % which are significantly lesser as compared to delete methods, with less compromise to the performance.

Qualitatively Fig 9, we can observe that for specific 3D volume the no delete methods show much less uncertainty between the boundary of the pancreas and the tumor (indicated by green arrows) while the delete methods are highly uncertain (indicated by red arrows).

On observation of representative segmentation for different active learning methods (Fig. 10) specifically for the healthy pancreas class it can be noticed that all methods perform well and are qualitatively similar. However, the proposed method Epistemic+MI+NoDelete utilizes a lesser amount of data to achieve a similar result (Tab. 2).

### 6.5. Pancreas Ablation Studies

For the pancreas dataset it can be observed that ensemble based approaches perform better than SVGD (Fig 11). It should also be noted that the combinations where frequency of hard-samples is increased (NoDelete) the performance is limited to 87% when compared with (Delete) combinations. However, the performance is achieved with a relatively lesser amount of data ( 50% or less).

### 6.6. Reduction of Uncertainty

Quantifying the uncertainty of the training pool for different methods of progressing active iterations Fig. 12. The top row is the uncertainty of all 3D voxels of all 3D volumes in the training pool of hippocampus dataset per active iteration. We can see how the distribution of uncertainty shifts towards lower values. Specifically, epistemic combined with MI without deleting points from unlabeled pool indicates the highest uncertainties at $5^{th}$ iteration as compared to all other methods. At $20^{th}$ iteration it is equivalent to all other methods and least at $25^{th}$ active iteration. A similar observation can be made about the pancreas dataset (bottom row).

Qualitatively we also show a TSNE [27] embedded clustering of the entire dataset based on the actual input images shown in Fig. 13. The first 6 components of a principal component analysis were used as features for TSNE embedding. The top row depicts the best active iterations of the three variants epistemic delete, epistemic + MI + delete and epistemic + MI + NoDelete and it can be seen that the proposed approach uses the least data and shows a relatively similar test Dice's score as other approaches. It should also be noted that the actively selected data points are located at the edges of the clusters and number of data points per data class seem to be relatively balanced. A similar observation can be made for the Pancreas dataset.

## 7. Discussion & Conclusion

### 7.1. Discussion

Duplicating already labeled images during active learning works well for the hippocampus dataset. We achieve a mean Dice score of 0.72 across 5 seeds (maximum achieved is 0.74) with 22% of data by utilizing the active learning methodologies. The fully supervised baseline by using all the data with an Ensemble and SVGD were found to be at 0.7257 and 0.7222. In comparison, the state-of-the-art performance is 0.86 on the MSD's challenge test set from [17], [18].

We would like to highlight the fact that the performance is in the relative ballpark to the state-of-the-art while it should be acknowledged that we used a lighter version of the U-net on downsampled images to allow for rapid experimentation. Specific improvements to the pre-processing pipeline such as ideal patch size, image resolution, patch sampling strategy and image resampling could lead to higher performance of the proposed active learning pipelines. Our specific pre-processing settings for each task might not be by no means optimal to achieve the best possible performance. Additionally, no data augmentation techniques were applied as we purely wanted to evaluate the benefit of active learning and hard-case selection for the labeled training dataset.

It should be noted that our study shows that the proposed methodologies work well on one dataset (hippocampus) (Fig. 7) but not as well on the other (pancreas) (Fig. 11). We therefore conclude that active learning methodologies are task and data dependent and the selection of different acquisition schemes needs to be evaluated in order to achieve acceptable performance.

On the other hand, the performance gains through active learning were limited for the pancreas dataset. This observation leads us back to address the problem of uncertainty, which is specific to high dimensionality data with low sample size. Notably, for the hippocampus dataset, the entire 3D volume was fed as input to the network which is not the case for the pancreas dataset (patches of $48 \times 48 \times 48$ were used) as the volumes ranged between 41 - 131 slices of sizes $77 \times 77$ - $125 \times 125$. It is also likely that the uncertainty estimates along organ boundaries could be insufficient for active selection; aggravated by the fact that the pancreas is a relatively small organ within the abdomen. Our methods focused on the active selection of entire image volumes. For small organ segmentation, like that of the pancreas, it is possible that more granular active patch selection strategies might perform better. It is possible that targeted mutual information of the fore-ground (specific to the organ) as compared to entire 3D volumes could be more helpful for high dimensional and low sample size datasets. This result indicates that active learning strategies often do not function well in all scenarios and highly depend on the dataset used, as shown in [15].

We also acknowledge as a limitation that model calibration has not been used as a part of our current work, but it is expected that standard calibration techniques such as T-scaling should lead to a more reliable uncertainty estimation [14]. There are multiple possibilities to improve the sequential process of active learning: 1.) Fine-tuning the same model on adding data rather than training from scratch again. Prior work for active fine-tuning [48] has been evaluated for biomedical classification and detection tasks. The effort indicated that fine-tuning reduces training time for some datasets but not necessarily for all. 2.) Using spatial dropout or M-heads at the end of the network to reduce the computational cost for both training and uncertainty estimates [20]. Although multiple works have shown that spatial dropout can be used for approximating the Bayesian inference needed for active learning, we did not find it suitable for the datasets used in this study.

A current limitation of the proposed active learning query-by-committee methods are their computational cost. The total training time for a single active learning method with 40 active iterations was approximately 160 and 60 GPU hours for pancreas and hippocampus datasets, respectively, for the methods explored in this study. As a future direction, it would be interesting to evaluate the efficiency of the methods by increasing the query size to reduce the active learning algorithm compute time. However, there is also the caveat of more annotation being required due to increased query size. In realistic conditions the GPU compute time would also depend upon the annotation resources available at hand and the size of the dataset.

### 7.2. Conclusion

We have shown that the proposed active learning framework can leverage highly uncertain data points based on model uncertainty by increasing their frequency in the training pool. Doing so can reduce the annotating cost for using deep learning to build medical imaging segmentation models.

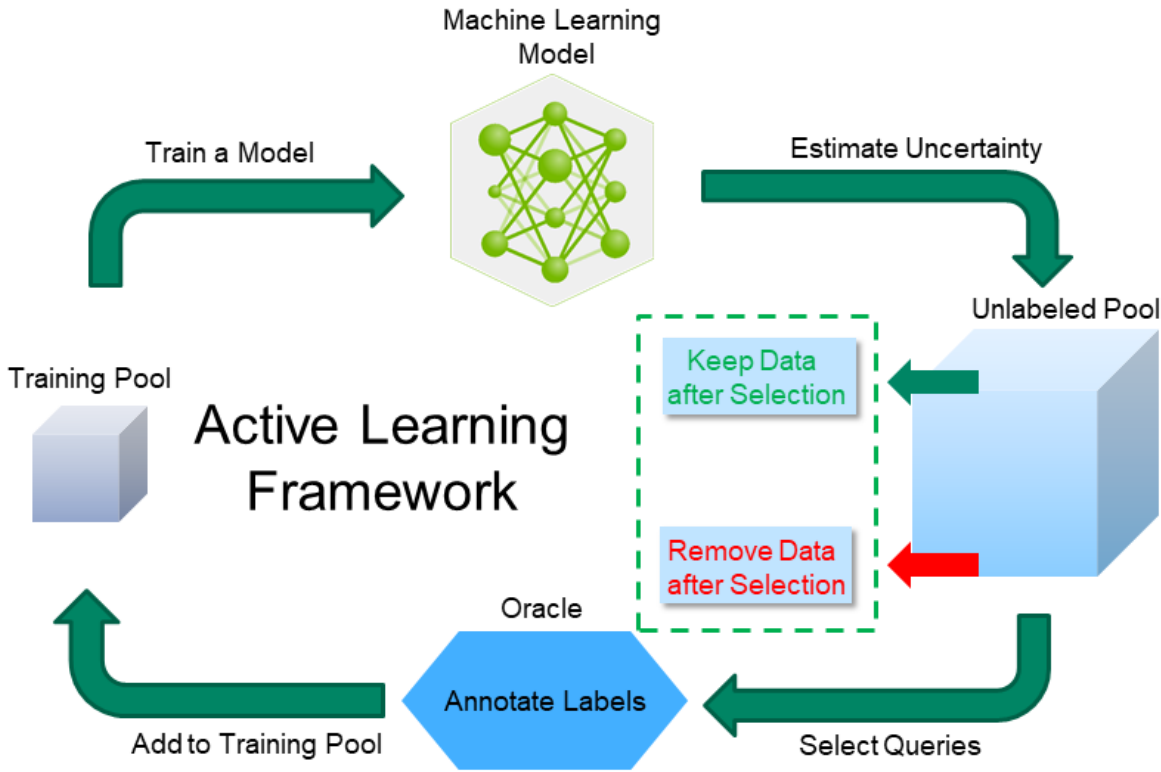

Figure 1: Proposed active learning framework. The data is actively selected from the unlabeled pool, annotated by oracle and then added to the training pool for model update.

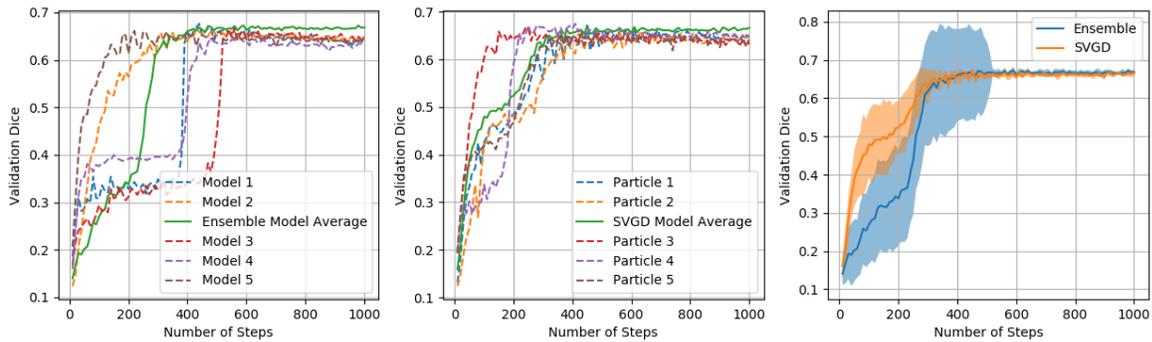

Figure 2: Validation Dice's scores for the Hippocampus dataset (30 Random data points were used for training). Left: models part of an ensemble and their average, Center: particles of SVGD and their average, Right: Variance across the members of ensemble and SVGD framework

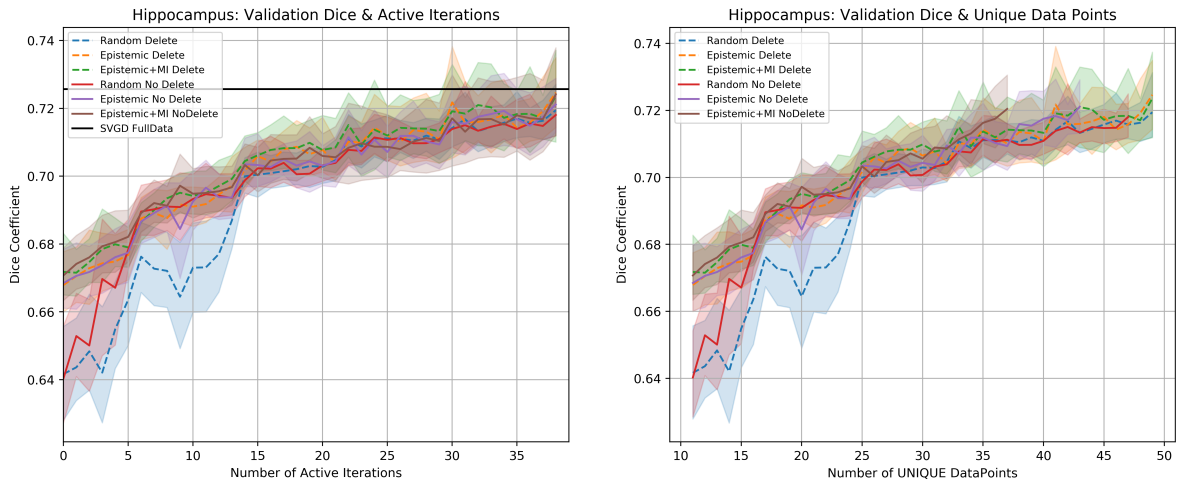

Figure 3: Quantitative results of Hippocampus dataset: Left) Mean Dice's scores and 90% confidence intervals of all volumes of all classes per active iteration for the validation set. Right) Mean Dice's scores of all volumes of all classes for the validation set with unique data points detected in training pool per active iteration. Total data points in unlabeled pool 153.

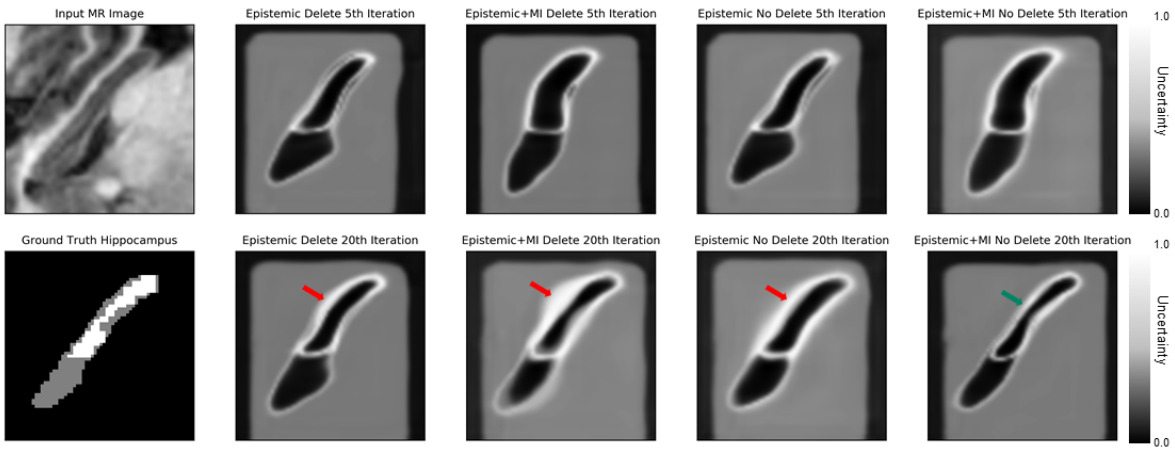

Figure 4: A 3D MRI volume selected by all acquisition methods for annotation. The top row shows the input image and the epistemic uncertainty at the 5th active iteration. The bottom row shows ground truth and uncertainty at 20th active iteration. The green arrow indicate lesser uncertainty detected by proposed method where data is not deleted from the unlabeled pool as compared to other methods.

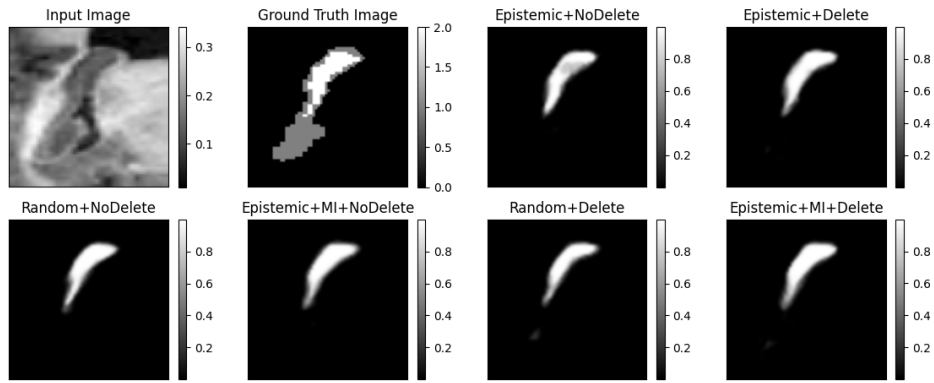

Figure 5: Top row from left to right shows the input image and the ground truth image. The successive ones are the prediction maps from the different methods for the second class of the hippocampus dataset. The model from the 37th active iteration was utilized from each method. It should be noted that all methods perform relatively similar and reasonable segmentation. However, the proposed methods Epistemic+MI+NoDelete utilize lesser data.

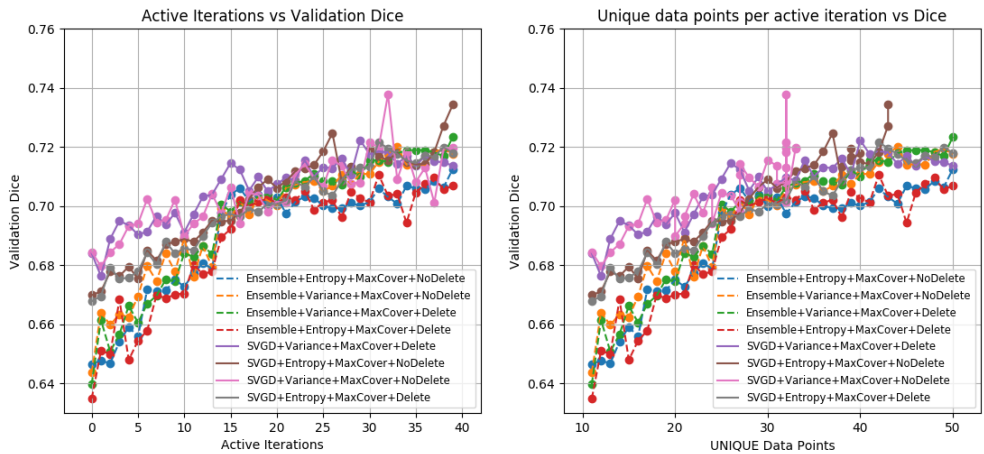

Figure 6: Left: Validation Dice's scores versus active iterations for maximum cover-based acquisition functions for SVGD and ensemble methods. Right: Mean Dice's scores of all volumes of all classes for the validation set with unique data points detected in the training pool per active iteration for maximum cover-based acquisition functions. The active learning committee of the ensemble is from prior work [26] and maximum-cover [13].

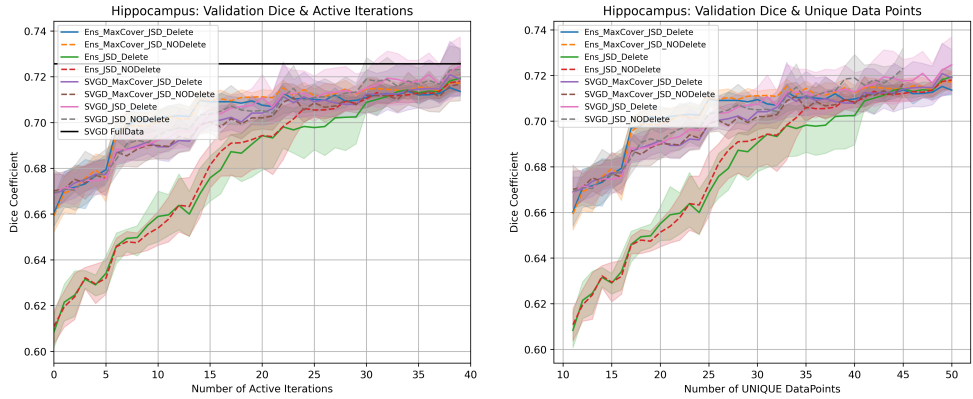

Figure 7: Left: Validation Dice's scores versus active iterations for maximum cover-based acquisition functions in combination with Jenson-Shannon divergence for SVGD and ensemble methods Right: Mean Dice's scores of all volumes of all classes for the validation set with unique data points detected in training pool per active iteration for maximum cover-based acquisition functions in combination with SVGD and ensemble methods. The active learning committee of the ensemble is from prior work [26], maximum-cover [13], Jenson-Shannon divergence [29]. NODelete is indicative of the usage of increasing frequency of hard-samples in the labeled pool of data and Delete implies vice-versa.

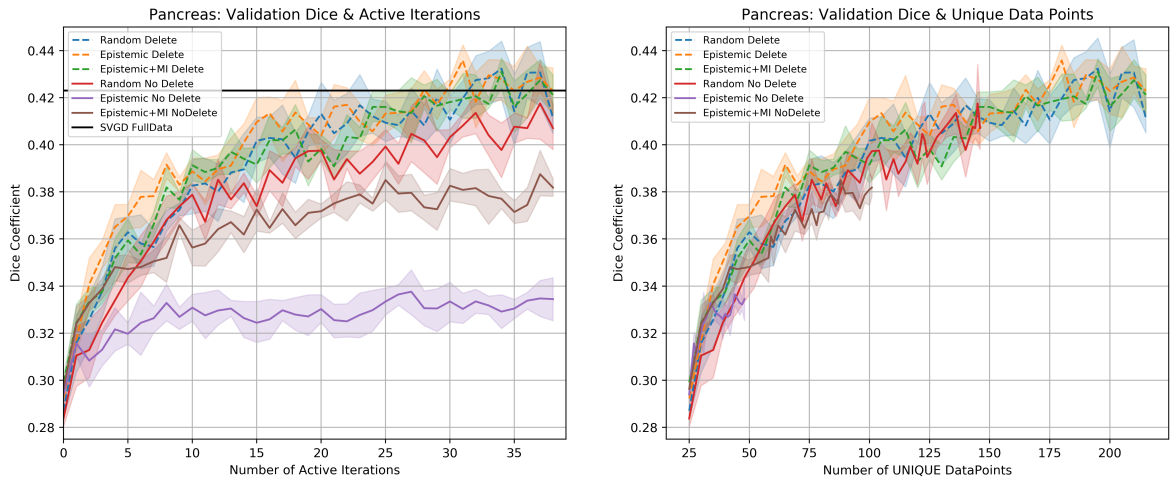

Figure 8: Quantitative results of Pancreas dataset: Left) Mean Dice's scores and 90% confidence intervals of all volumes of all classes per active iteration for the validation set. Right) Mean Dice's scores of all volumes of all classes for the validation set with unique data points detected in training pool per active iteration. Total amount of data points in unlabeled pool is 201.

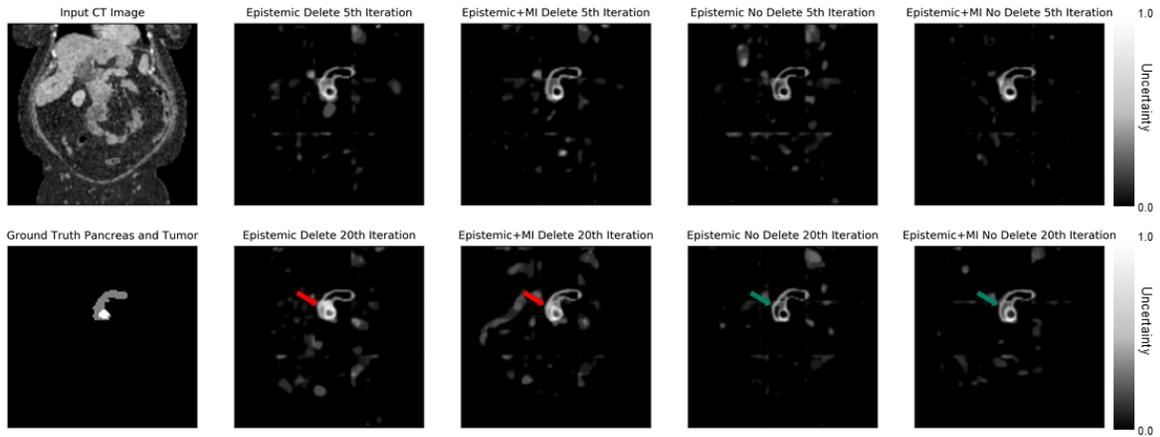

Figure 9: A 3D CT volume selected by all acquisition methods for annotation. The top row shows the input image and the epistemic uncertainty at $5^{th}$ active iteration. The bottom row shows ground truth and uncertainty at $20^{th}$ active iteration. The green arrows indicate lesser uncertainty detected by methods where data is not deleted from unlabeled pool vice-versa for red arrows. It should be noted that the CT volume was down-sampled to $4\ mm$ isotropic resolution for this experiment.

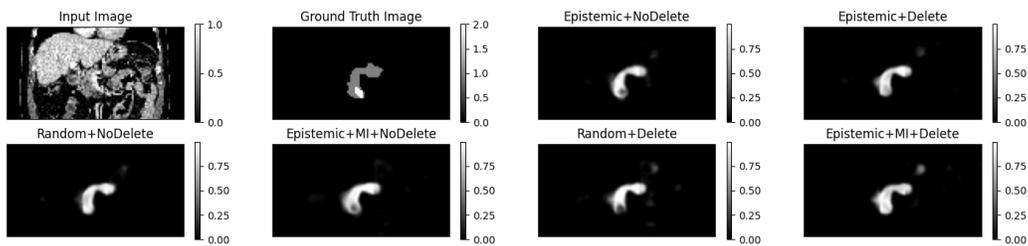

Figure 10: Top row from left to right shows the input image and the ground truth image. The successive ones are the prediction maps from the different methods of the healthy pancreas tissue. The model from the 37th active iteration was utilized from each method. It should be noted that all methods perform relatively similarly and achieve reasonable segmentations. However, the proposed method Epistemic+MI+NoDelete utilizes lesser data as shown in Table II.

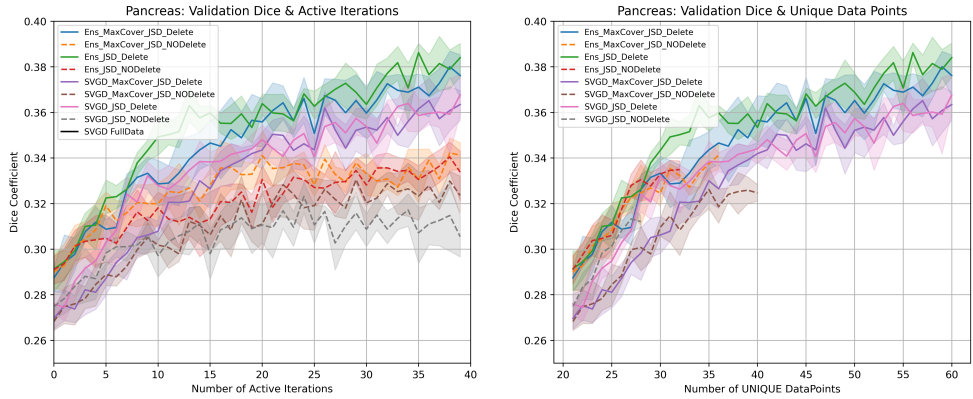

Figure 11: Left: Validation Dice's scores versus active iterations for maximum cover-based acquisition functions in combination with Jenson-Shannon divergence for SVGD and ensemble methods Right: Mean Dice's scores of all volumes of all classes for the validation set with unique data points detected in training pool per active iteration for maximum cover-based acquisition functions in combination with SVGD and ensemble methods. The active learning committee of the ensemble is from prior work [26], maximum-cover [13], Jenson-Shannon divergence [29]. NODelete is indicative of the usage of increasing frequency of hard-samples in the labeled pool of data and Delete implies vice-versa

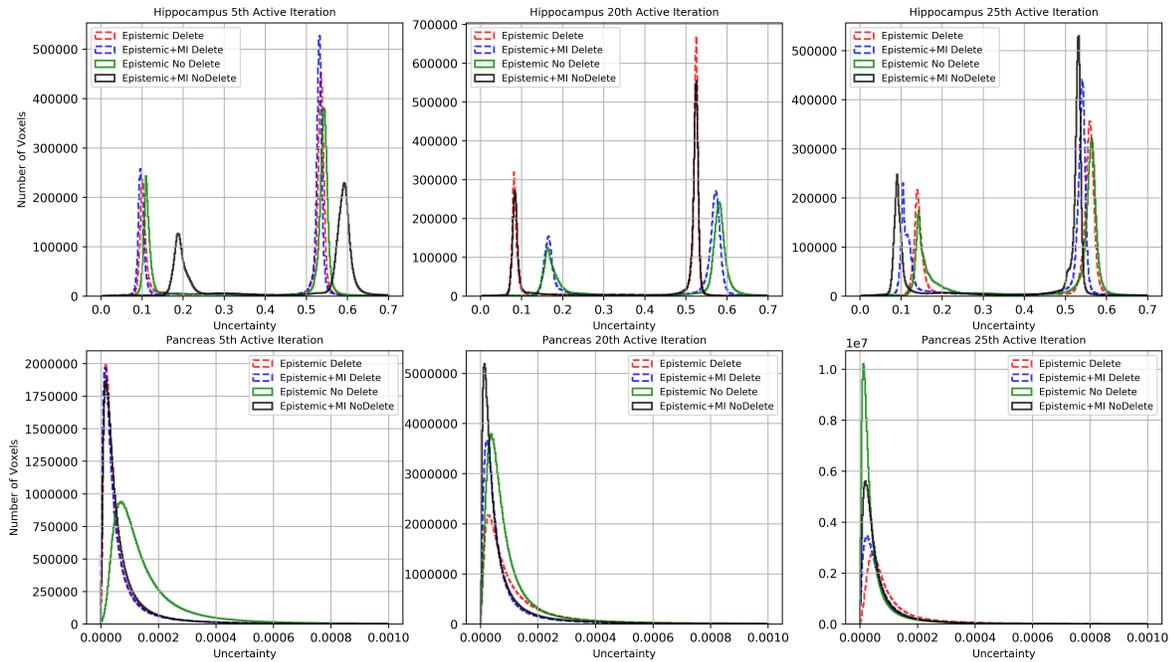

Figure 12: Top row depicts distribution of uncertainty for training pool of hippocampus dataset for $5^{th}$, $20^{th}$, and $25^{th}$ active iteration respectively. Bottom row indicates the same as above for the pancreas dataset.

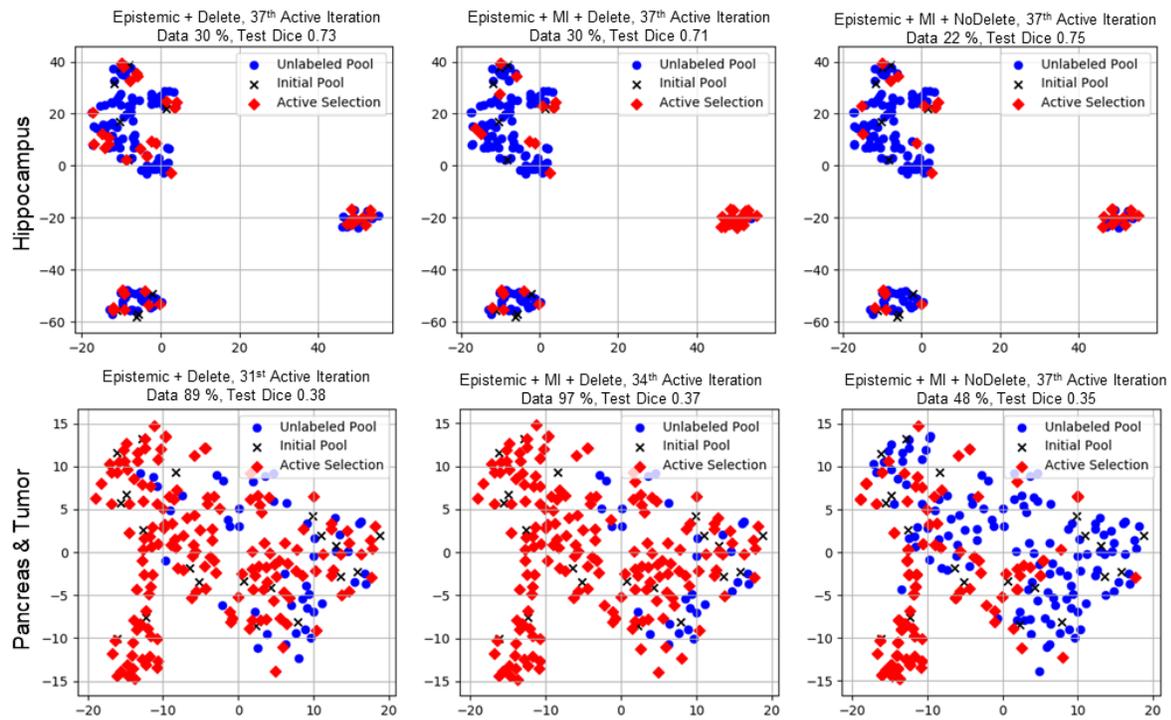

Figure 13: TSNE embedded clustering of the entire dataset where the actively selected data points, initial points and the unlabeled pool are shown for the best active iterations for both datasets of Hippocampus (Top Row) and Pancreas (Bottom Row).